\def\year{2019}\relax
\documentclass[letterpaper]{article} 
\usepackage{aaai19}  
\usepackage{times}  
\usepackage{latexsym}
\usepackage{makecell,multirow,diagbox,tabu}
\usepackage{amsmath}
\usepackage{amssymb}
\usepackage{booktabs}
\usepackage{subfigure}
\usepackage{bm}
\usepackage{soul}
\usepackage{color}
\usepackage{colortbl}
\usepackage{helvet}  
\usepackage{courier}  
\usepackage{url}  
\usepackage{graphicx}  
\begin{document}
%
\title{Exploiting Coarse-to-Fine Task Transfer for \\ Aspect-level Sentiment Classification}

\author{Zheng Li$^{\dag}$, Ying Wei$^{\star}$, Yu Zhang$^{\dag}$, Xiang Zhang$^{\dag}$, Xin Li$^{\pm}$, Qiang Yang$^{\dag}$\\
$^{\dag}$Hong Kong University of Science and Technology, Hong Kong\\
$^{\star}$Tencent AI Lab\\
$^{\pm}$The Chinese University of Hong Kong, Hong Kong\\
\{zlict,qyang\}@cse.ust.hk, judywei@tencent.com, yu.zhang.ust@gmail.com, xzhangax@ust.hk, lixin@se.cuhk.edu.hk}
\maketitle

\begin{abstract}
Aspect-level sentiment classification (ASC) aims at identifying sentiment polarities towards aspects in a sentence, where the aspect can behave as a general {\it Aspect Category} (AC) or a specific {\it Aspect Term} (AT). However, due to the especially expensive and labor-intensive labeling, existing public corpora in AT-level are all relatively small. Meanwhile, most of the previous methods rely on complicated structures with given scarce data, which largely limits the efficacy of the neural models. In this paper, we exploit a new direction named {\it coarse-to-fine task transfer}, which aims to leverage knowledge learned from a rich-resource source domain of the coarse-grained AC task, which is more easily accessible, to improve the learning in a low-resource target domain of the fine-grained AT task. To resolve both the aspect granularity inconsistency and feature mismatch between domains, we propose a Multi-Granularity Alignment Network (MGAN). In MGAN, a novel Coarse2Fine attention guided by an auxiliary task can help the AC task modeling at the same fine-grained level with the AT task. To alleviate the feature false alignment, a contrastive feature alignment method is adopted to align aspect-specific feature representations semantically. In addition, a large-scale multi-domain dataset for the AC task is provided. Empirically, extensive experiments demonstrate the effectiveness of the MGAN. 
\vspace{-2.5mm}
\end{abstract}

\section{Introduction}
Aspect-level sentiment classification (ASC) aims to infer sentiment polarities over aspect categories (AC) or aspect terms (AT) distributed in sentences~\cite{pang2008opinion,liu2012sentiment}. An aspect category implicitly appears in the sentence, which describes a general category of the entities. For example, in the sentence ``{\it The salmon is tasty while the waiter is very rude}'', the user speaks positively and negatively towards two aspect categories {\it ``{\textbf{food}}''} and {\it ``{\textbf{service}}''}, respectively. An aspect term characterizes a specific entity that explicitly occurs in the sentence. Considering the same sentence ``{\it The \textbf{salmon} is tasty while the \textbf{waiter} is very rude}'', the aspect terms are ``{\it \textbf{salmon}}'' and ``{\it \textbf{waiter}}'', and the user expresses positive and negative sentiments over them, respectively. In terms of the aspect granularity, the AC task is coarse-grained while the AT task is fine-grained.

To model aspect-oriented sentiment analysis, equipping Recurrent Neural Networks (RNNs) with the attention mechanism has become a mainstream approach~\cite{tang2015effective,wang2016attention,ma2017interactive,chen2017recurrent}, where RNNs aim to capture sequential patterns and the attention mechanism is to emphasize appropriate context features for encoding aspect-specific representations. Typically, attention-based RNN models can achieve good performance only when large corpora are available. 
However, AT-level datasets require the aspect terms to be comprehensively manually labeled or extracted by sequence labeling algorithms from the sentences, which is especially costly to obtain. Thus, existing public AT-level datasets are all relatively small, which limits the potential of neural models. 

Nonetheless, we observe that plentiful AC-level corpora are more easily accessible. This is because that aspect categories are usually in a small set of general aspects that can be pre-defined. For example, commercial services such as review sites or social media can define a set of valuable aspect categories towards products or events in a particular domain (e.g., ``{\it food}'', ``{\it service}'', ``{\it speed}'', and ``{\it price}'' in the {\it Restaurant} domain). As a result, the mass collections of user preferences towards different aspect categories become practicable. Motivated by this observation, we propose a new problem named~\emph{coarse-to-fine task transfer} across both domain and granularity, with the aim of borrowing knowledge from an abundant source domain of the coarse-grained AC task to a small-scale target domain of the fine-grained AT task.

The challenges in fulfillment of this setting are two-fold: (1) task discrepancy: the two tasks concern with the aspects with different granularity. Source aspects are coarse-grained aspect categories, which lack a priori position information in the context. However, target aspects are fine-grained aspect terms, which have accurate position information. Thus, inconsistent granularity in aspects causes the discrepancy between tasks; (2) feature distribution discrepancy: generally the domains in the two tasks are different, which causes the distribution shift for both the aspects and its context between domains. For example, in the source {\it Restaurant} domain, {\it tasty} and {\it delicious} are used to express positive sentiment towards the aspect category ``{\it food}'', while {\it lightweight} and {\it responsive} often indicate positive sentiment towards the aspect term ``{\it mouse}'' in the target {\it Laptop} domain. 

To resolve the challenges, we propose a novel framework named \textbf{M}ulti-\textbf{G}ranularity \textbf{A}lignment \textbf{N}etwork (MGAN) to simultaneously align aspect granularity and aspect-specific feature representations across domains. Specifically, the MGAN consists of two networks for learning aspect-specific representations for the two domains, respectively. First, to reduce the task discrepancy between domains, i.e., modeling the two tasks at the same fine-grained level, we propose a novel Coarse2Fine (C2F) attention module to help the source task automatically capture the corresponding aspect term in the context towards the given aspect category (e.g., ``salmon'' to the ``{\it food}''). Without any additional labeling, the C2F attention module can learn the coarse-to-fine process by an auxiliary task. Actually, more specific aspect terms and their position information are most directly pertinent to the expression of sentiment. The C2F module makes up these missing information for the source task, which effectively reduces the aspect granularity gap between tasks and facilitates the subsequent feature alignment. 

Second, considering that a sentence may contain multiple aspects with different sentiments, thus capturing incorrect sentiment features towards the aspect can mislead feature alignment. To prevent false alignment, we adopt the Contrastive Feature Alignment (CFA)~\cite{motiian2017unified} to semantically align aspect-specific representations. The CFA considers both semantic alignment by maximally ensuring the equivalent distributions from different domains but the same class, and semantic separation by guaranteeing distributions from both different classes and domains to be as dissimilar as possible. Moreover, we build a large-scale multi-domain dataset named {\emph{YelpAspect}} with 100K samples for each domain to serve as highly beneficial source domains. Empirically, extensive experiments demonstrate that the proposed MGAN model can achieve superior performances on two AT-level datasets from SemEval`14 ABSA challenge and an ungrammatical AT-level twitter dataset.





Our contributions of this paper are four-fold: (1) to the best of our knowledge, a novel transfer setting cross both domain and granularity is first proposed for aspect-level sentiment analysis; (2) a new large-scale, multi-domain AC-level dataset is constructed; (3) the novel Coarse2Fine attention is proposed to effectively reduce the aspect granularity gap between tasks; (4) empirical studies verify the effectiveness of the proposed model on three AT-level benchmarks.

\vspace{-1mm}
\section{Related Work}
Traditional supervised learning algorithms highly depend on extensive handcrafted features to solve aspect-level sentiment classification~\cite{jiang2011target,kiritchenko2014nrc}. These models fail to capture semantic relatedness between the aspect and its context. To overcome this issue, the attention mechanism, which has been successfully applied in many NLP tasks~\cite{bahdanau2014neural,sukhbaatar2015end,yang2016hierarchical,shen2017disan}, can help the model explicitly capture intrinsic aspect-context association~\cite{tang2015effective,tang2016aspect,wang2016attention,ma2017interactive,chen2017recurrent,ma2018targeted,li2018transformation}. However, most of these methods highly rely on data-driven RNNs or tailor-made structures to deal with complicated cases, which requires substantial AT-level data to train effective neural models. Different from them, the proposed model can highly benefit from useful knowledge learned from a related abundant domain of the AC-level task. 

Existing domain adaptation tasks for sentiment analysis focus on traditional sentiment classification without considering the aspect~\cite{blitzer2007biographies,pan2010cross,glorot2011domain,chen2012marginalized,bollegala2013cross,yu2016learning,li2017end,li2018hatn}. In terms of data scarcity and the value of task, transfer learning is more urgent for aspect-level sentiment analysis that characterizes users` different preferences. To the best of our knowledge, only a few studies have explored to transfer from a single aspect category to another in a same domain based on adversarial training~\cite{zhang2017aspect}. Different from that, we explore a motivated and challenging setting which aims to transfer cross both aspect granularity and domain. 

\begin{figure*}[thb!]
\centering
\includegraphics[width=0.81\linewidth]{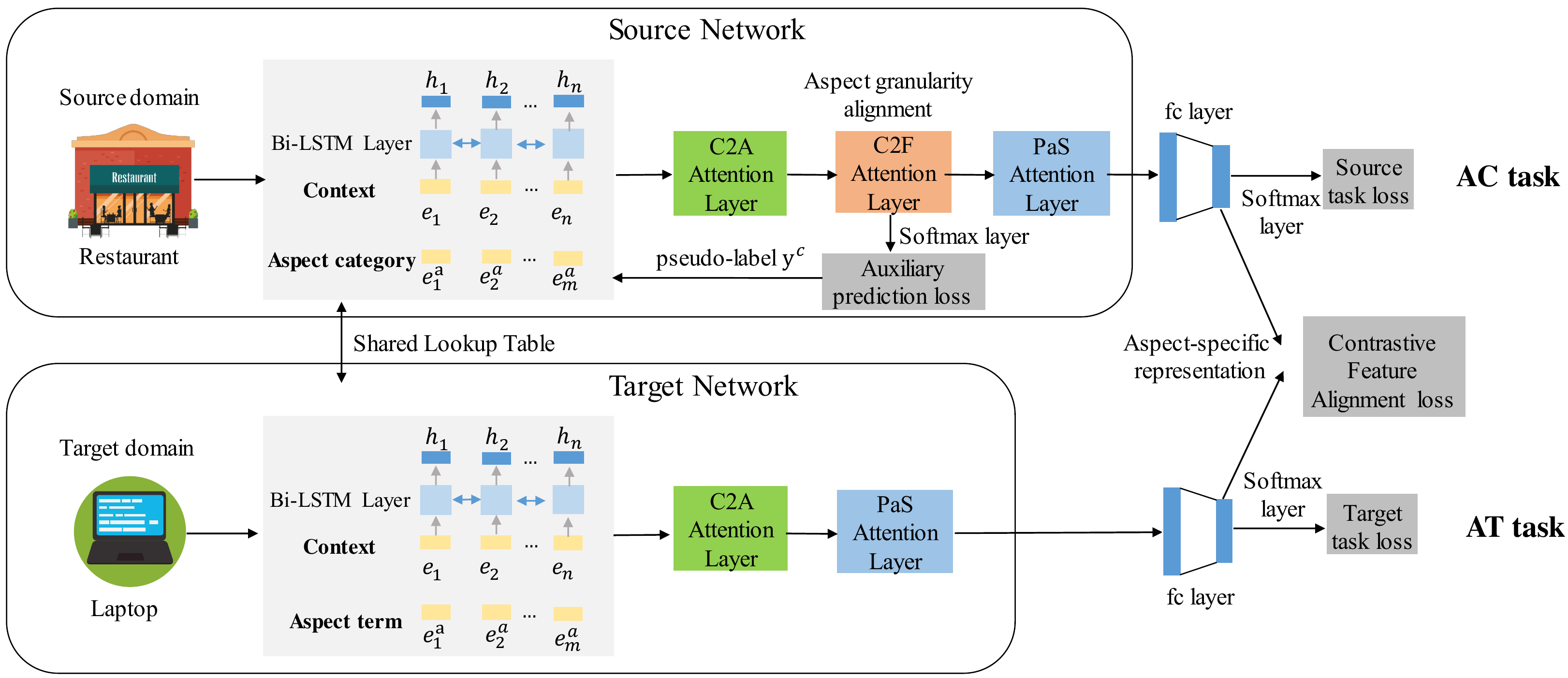}
\vspace{-3mm}
\caption{The architecture of the Multi-Granularity Alignment Network (MGAN) model.}
\vspace{-4mm}
\label{fig:overall}
\end{figure*}

\section{Multi-Granularity Alignment Network} 
In this section, we introduce the proposed MGAN model. We first present the problem definition and notations, followed by an overview of the model. Then we detail the model with each components.

\vspace{-0.5mm}
\subsection{Problem Definition and Notations}
\noindent \textbf{Coarse-to-fine task transfer}
Suppose that we have sufficient AC-level labeled data $\small{{ \mathbf{X} }^{ s }\!=\!{ \{ ({ \mathbf{x} }_{k }^{ s },{ { \mathbf{a} } }_{ k }^{ s }),{ y }_{ k }^{ s } \}  }_{ k\!=\!1 }^{ { N }^{ s } }}$ in a source domain $D_{s}$, where $\small{y^s_k}$ is the sentiment  label for the $k$-th sentence-aspect pair  $(\small{\mathbf{x}_k^s},\small{\mathbf{a}_k^s})$. Besides, only a small amount of AT-level labeled data $\small{{ \mathbf{X} }^{ t }\!=\!{ \{ ({ \mathbf{x} }_{ k' }^{ t },{ { \mathbf{a} } }_{ k' }^{ t }),{ y }_{ k' }^{ t } \}  }_{ k'\!=\!1 }^{ { N }^{ t } }}$ is available in a target domain $D_{t}$. Note that each source aspect $\small{\mathbf{a}^s_{k}}$ belongs to a set of pre-defined aspect categories $C$ while each target aspect $\small{\mathbf{a}^t_{k'}}$ is a sub-sequence of $\small{\mathbf{x}^t_{k'}}$, i.e., aspect term. The goal of this task is to learn an accurate classifier to predict the sentiment polarity of target testing data. 


\subsection{An Overview of the MGAN model}
The goal of the MGAN aims to transfer from a rich-resource source domain of an AC task to facilitate a low-resource target domain of an AT task. The architecture of the proposed MGAN is shown in Figure~\ref{fig:overall}. Specifically, the MGAN consists of two networks for tackling the two aspect-level tasks respectively. To reduce the task discrepancy, the two networks contain different numbers of attention hops such that they can keep a consistent granularity and the symmetric information towards the aspect. In MGAN, two basic hop units are used similarly as common attention-based RNN models, where the Context2Aspect (\textbf{C2A}) attention aims to measure the importance of each aspect word and generate the aspect representation with the aid of each context word, and the Position-aware Sentiment (\textbf{PaS}) attention utilizes the obtained aspect representation and the position information of the aspect to capture relevant sentiment features in the context for encoding the aspect-specific representation. 

Moreover, we build a Coarse2Fine (\textbf{C2F}) attention upon the C2A module to specifically model the source aspect before feeding to the PaS module. The C2F module uses the source aspect representation to attend corresponding aspect terms in the context and then the attended context features is conversely predicted the category of the source aspect (pseudo-label). After obtaining aspect-specific representations, the knowledge transfer between the two tasks is via the contrastive feature alignment. In summary, the source network acts as a ``teacher'', which consists of three-level attention hops (C2A+C2F+PaS) for the AC task, while the target network is like a ``student'' that only uses two basic attention hops (C2A+PaS) for the AT task. In the following sections, we introduce each component of MGAN in details.


\subsection{Bi-directional LSTM layer}
Given a sentence-aspect pair ($\mathbf{x}$, $\mathbf{a}$) from the source or target domain, we assume that the sentence consists of $n$ words, i.e., $\small{\mathbf{x}\!=\!\{w_{1},w_{2},...,w_{n}\}}$, and the aspect contains $m$ words, i.e., $\small{\mathbf{a}\!=\!\{w^{a}_{1},w^{a}_{2},...,w^{a}_{m}\}}$.
Then we map them into its embedding vectors $\small{{ \mathbf{e} }\!=\!{ \{ \mathbf{e}_{ i } \}  }_{ i=1 }^{ n } \!\in\! \mathbb{R}^{n\!\times\! d_{w}}}$ and $\small{{ \mathbf{e}^{a} }\!=\!{ \{ \mathbf{e}^{a}_{ j } \}  }_{ j=1 }^{ m } \!\in\! \mathbb{R}^{m\!\times\! d_{w}}}$ respectively. To capture phrase-level sentiment features in the context (e.g., ``not satisfactory''), we employ a Bi-directional LSTM (Bi-LSTM) to preserve the contextual information for each word of the input sentence. The Bi-LSTM transforms the input $\small{{ \mathbf{e} }}$ into the contextualized word representations $\small{{ \mathbf{h} }\!=\!{ \{ \mathbf{h}_{ i } \}  }_{ i=1 }^{ n } \!\in\! \mathbb{R}^{n\!\times\! 2d_{h}}}$ (i.e. hidden states of Bi-LSTM). For simplicity, we denote the operation of an LSTM unit on ${ \mathbf{e} }_{ i }$ as LSTM$( { \mathbf{e} }_{ i } )$. Thus, the contextualized word representation $\small{{ \mathbf{h} }_{ i }\!\in\!\mathbb{R}^{2d_h}}$ is obtained as
\begin{equation}
{ \mathbf{h} }_{ i }= [ \overrightarrow { { \textmd{LSTM} } } ( { \mathbf{e} }_{ i } ) ;\overleftarrow { \textmd{LSTM} } ( { \mathbf{e} }_{ i } )  ] , i\!\in\! [ 1,n ],
\end{equation}
where ``$;$'' denotes the vector concatenation.

\subsection{Context2Aspect (C2A) Attention}
\label{catr}
Not all aspect words contribute equally to the semantic of the aspect. For example, in the aspect term ``{\it techs at HP}'', the sentiment is usually expressed over the headword ``{\it techs}'' but seldom over modifiers like the brand name ``{\it HP}''. Thus, ``{\it techs}'' is more important than ``{\it at}'' and ``{\it HP}''. This also applies to the aspect category (e.g., ``{\it food seafood fish}''). Thus, we propose the Context2Aspect (C2A) attention to measure the importance of the aspect words with regards to each context word. Formally, we calculate a pair-wise alignment matrix $\mathbf{M}\!\in\! \mathbb{R}^{n \!\times\! m}$ between the context and the aspect, where the alignment score $M(i,j)$ between the $i$-th context word and the $j$-th aspect word is obtained as
\begin{equation}
M(i,j)=\mathrm{tanh}({ { { \mathbf{W} } } }_{a}[{ \mathbf{h} }_{ i};{ { \mathbf{e} } }^{a}_{ j }]+{ b }_{ a }),
\end{equation}
where $\small{{ \mathbf{W} }_{ a }}$ and $\small{{b}_{a}}$ are learnable parameters. Then, we apply a row-wise softmax function to get probability distributions in each row. By defining ${ \bm{\delta }  }(i)\!\in\!\mathbb{R}^{m}$ as the individual aspect-level attention given the $i$-th context word, we average all the ${ \bm{\delta }  }(i)$'s to get the C2A attention as $\small{\bm{{\alpha}}\!=\!\frac { 1 }{ n } \sum _{ i=1 }^{n }{ \bm{\delta }(i)} } $. The C2A attention further contributes the context-aware aspect representation by ${ \mathbf{h} }^{ a }_{*}\!=\!\sum _{ j=1 }^{ m }{ { { {\alpha}  }_{j} \mathbf{e} }_{ j }^{ a  } }$, where $\small{*\!\in\!\{s,t\}}$ denotes the source or target domain. We tackle the aspect representation ${ \mathbf{h} }^{ a }_{*}$ for the two tasks differently, where ${ \mathbf{h} }^{ a }_{s}$ is fed to the C2F module while ${ \mathbf{h} }^{ a }_{t}$ is directly fed to the PaS module.


\subsection{Coarse2Fine (C2F) Attention}
Aspect terms, which act as the true ``opinion entity'', are the most directly pertinent to the expression of sentiment. However, source task concerns with coarse-grained aspect categories that lack of detailed position information in the context. We wish to achieve task alignment such that the target task can leverage more useful knowledge learned from the source task at the same fine-grained level. It is observed that the number of source aspects is much smaller and many instances contain same aspect category, but the underlying entities can behave diversely in different contexts. For example, the aspect category {\it ``food seafood fish''} can be instantiated as ``salmon'', ``tuna'', ``taste'' and etc. 

Based on this observation, we can capture more specific semantics of the source aspect and its position information conditioned on its context. Motivated by autoencoders~\cite{bengio2007greedy}, we introduce an auxiliary pseudo-label prediction task for the source task. In this task, a source aspect $\small{\mathbf{a}^s}$ is not only regarded as a sequence of aspect words, but also as a pseudo-label (category of the aspect) $\small{{y}^c}$, where $c \in C$ and $C$ is a set of aspect categories. We utilize the obtained aspect representation ${ \mathbf{h} }^{ a }_{s}$ for $\small{\mathbf{a}^s}$ to attend the context and then the induced attention scores aggregate the context information to conversely predict the pseudo category label of $\small{\mathbf{a}^s}$ itself. If the context contains the aspect term correlated closely to the source aspect, then the attention mechanism can emphasize it for better prediction. We denote this mechanism as Coarse2Fine attention, which is calculated as:
\begin{eqnarray}
&&\small{{ z }_{ i }^{ f }\!=\!{ ({ \mathbf{u} }_f) }^{ T }\mathrm{tanh}({ { { { { \mathbf{W} } } } }_f }[{ \mathbf{h} }_{ i };{ { { \mathbf{h} } }^{ a  }_{s} }]+{ { \mathbf{b} } }_{ f })},\\ 
&&\small{{ \beta   }_{ i }^{ f }\!=\!\frac { {\exp}({ z }_{i }^{ f }) }{ \sum _{ i'=1 }^{ n }{ {\exp}({ z }_{i'}^{ f }) }  }},\\
&&{ \mathbf{v} }^{ a }=\sum _{ i=1 }^{ n }{ { \beta   }_{ i }^{ f }{ \mathbf{h} }_{ i } } ,
\end{eqnarray}
where $\small{{ \mathbf{W} }_{ f }\!\in\! { \mathbb{R} }^{ { d }_{ u }\!\times\! (2{ d }_{ h }\!+\!{ d }_{ e }) }}$, $\small{{ \mathbf{b} }_{ f } \!\in\! { \mathbb{R} }^{ { d }_{ u } }}$ and $\small{{ \mathbf{u} }_{ f } \!\in\! { \mathbb{R} }^{ { d }_{ u } }}$ are the weights of the layer. We feed the attended representation ${ \mathbf{v} }^{ a }$ to a {\it softmax} layer for the auxiliary task prediction, which is trained by minimizing the cross-entropy loss between the predicted pseudo-label $\hat{ y }^{c}_{ k }$ and its ground-truth ${  { y }  }^{c}_{ k }$ as:
\begin{equation}
\small{\mathcal{L}_{aux}=-\frac { 1 }{ { N }_{ s }} \sum _{ k=1 }^{ { N }_{ s } } \sum _{ c\in C }{ {  { y }  }^{c}_{ k }\log \hat{ y }^{c}_{ k } }.}
\end{equation}

However, there may not exist corresponding aspect term when the context implicitly expresses a sentiment toward the aspect category. To overcome this issue, similar to the gate mechanism in RNN variants~\cite{jozefowicz2015empirical}, we adopt a fusion gate $\mathbf{F}$ to adaptively controls the passed proportions of ${ \mathbf{h} }^{ a}_{s}$ and ${\mathbf{v} }^{a}$ towards a more specific source aspect representation ${ \mathbf{r} }^{ a }_{s}$: 
\begin{eqnarray}
&&\mathbf{F}=\mathrm{sigmoid} ({ \mathbf{W} }[{ \mathbf{v} }^{ a }; { \mathbf{h} }_{s}^{a }]\!+\!{ \mathbf{ b} }), \\
&&{ \mathbf{r} }^{ a }_{s}=\mathbf{F}\odot { \mathbf{h} }_{s}^{ a}+(\mathbf{1}- \mathbf{F})\odot {\mathbf{W'}}{ \mathbf{v} }^{ a },
\end{eqnarray}
where $\small{{ \mathbf{W} }\!\in\! {\mathbb{R}}^{ { d }_{ e } \!\times\! ({ d }_{ e }\!+\!2{ d }_{ h })}}$ and ${{ \mathbf{b} }\!\in\! {\mathbb{R}}^{ { d }_{ e }} }$ are the weights of the gate, $\small{{\mathbf{W'}} \!\in\! \mathbb{R}^{ { d }_{ e } \!\times\! { 2d }_{ h }}}$ performs dimension reduction, and $\odot$ denotes element-wise multiplication.

\subsection{Position-aware Sentiment (PaS) Attention}
Following an important observation found in~\cite{tang2016aspect,chen2017recurrent} that a closer sentiment word is more likely to be the actual modifier of the aspect term (e.g., in ``{\it great food but the service is dreadful}'', ``{\it great}'' is more closer to ``{\it food}'' than ``{\it service}''), we take the position information of the aspect term into consideration for designing the PaS attention. For the target domain, we adopt a proximity strategy to calculate the target position relevance between the $i$-th context word and aspect term as follows:
\begin{equation}
{ p }^{ t }_{ i } = \left\{ \begin{array}{ll}
1-\frac { m_0-i }{ n }         & i<m_0\\
0                                       & m_0\le i\le m_0+m\\
1-\frac { i-(m_0+m) }{ n } & i>m_0+m
\end{array} \right.,
\end{equation}
where $m_0$ is  the index of the first aspect word, $n$ and $m$ are the length of the sentence and aspect, respectively. 

Unfortunately, in the source domain where aspect category is given, the exact position of the corresponding aspect term is not directly accessible. Instead, the C2F attention vector ${\bm{ \beta  }}^{f } \!\in\! \mathbb{R}^{n}$, indicating the probability of each context word being an aspect term, can help establish the position relevance. We first define a location matrix $\small{\mathbf{L}\!\in\! \mathbb{R}^{{n}\!\times\!{n}}}$ to represent the proximity of each word in the sentence:
\begin{equation}
{ L }_{ ii'}\!=\!1-\frac { | i-i' |  }{ n }, i,i'\!\in\![1,n].\\
\end{equation}
Then we calculate the source position relevance for $i$-th context word with the aid of C2F attention weights by
\begin{equation}
{ p }^{ s }_{i}=\mathbf{L}_{i}{ \bm{\beta}  }^{f }.
\end{equation}
Obviously, the $i$-th context word closer to a possible aspect term with a large value in ${ \bm{\beta}  }^{ f }$ will have a larger position relevance ${ p }^{ s }_{i}$. Finally, the PaS attention is calculated by a general form for both domains: 
\begin{eqnarray}
&&\small{{ z }_{ i }^{ o }\!=\!{ ({ \mathbf{u} }_o) }^{ T }\mathrm{tanh}({ { { { { \mathbf{W} } } } }_o }[{ \mathbf{h} }_{ i };{ { { \mathbf{r}^{a}_{*} } }}]+{ { \mathbf{b} } }_o)},\\ 
&&\small{{ \gamma   }_{ i }^{ o }=\frac { { \exp }({ p }_{ i }^{*}{ z }_{ i}^{  o }) }{ \sum _{ i'=1 }^{ n }{ { \exp }({ p }_{ i'}^{*}{ z }_{i' }^{ o }) }  } },\\
&&\small{{ {\mathbf{v}}^{o} }=\sum _{ i=1 }^{ n }{ {  \gamma  }_{ i }^{ o }{\mathbf{h} }_{ i } }},
\end{eqnarray}
where ${ p }_{ i }^{*}$ is the position relevance and ${ \mathbf{r}^{a}_{*} }$ is the input aspect representation, with $\small{*\!\in\!\{s,t\}}$ denoting the source or target domain (note that ${ \mathbf{r}^{a}_{t} }=\mathbf{h}^{a}_{t}$). Then we pass the aspect-specific representation $\mathbf{v}^{o}$ to a fully-connected layer and a {\it softmax} layer for sentiment classification. The sentiment classification tasks for both domain are trained by minimizing two cross-entropy losses $\mathcal{L}_{sen}^{s}$ and $\mathcal{L}_{sen}^{t}$, respectively.


\subsection{Contrastive Feature Alignment}
After obtaining aspect-specific representations of two domains at the same granularity, we would further bridge the distribution gap across domains.
The prevalent unsupervised domain adaptation methods~\cite{gretton2007kernel,ganin2016domain} require enormous unlabeled target data to achieve satisfactory performances, which is impractical in our problem where collecting unlabeled data needs labor-intensive annotations of all aspect terms in the sentences. 
Therefore, inspired by~\cite{motiian2017unified}, we perform Contrastive Feature Alignment (CFA) by fully utilizing the limited target labeled data to semantically align representations across domains. Mathematically, we parameterize the two networks by ${g}_s$ and ${g}_t$, and denote the probability distribution by $\mathbb{P}$. Specifically, the CFA consisits of semantic alignment (SA) and semantic separation (SS). The SA aims to ensure identical distributions of feature representations $\mathbb{P}(g_s({ \mathbf{X} }^{ s }))$ and $\mathbb{P}(g_t({ \mathbf{X} }^{ t }))$ conditioned on different domains but the same class,  while the SS further alleviates false alignment by guaranteeing $\mathbb{P}(g_s({ \mathbf{X} }^{ s }))$ and $\mathbb{P}(g_t({ \mathbf{X} }^{ t }))$ to be as dissimilar as possible conditioned on both different domains and classes. Considering that only a small amount of target labeled data is available, we revert the CFA characterizing distributions with enough data to pair-wise surrogates as: 
\begin{equation}
\label{cfa}
\small{{{{{ \mathcal{L} }_{ cfa }\!=\!\sum _{ k,k' }^{  }{ \omega({  g_s({ \mathbf{x} }_{ k }^{ s },{ \mathbf{a} }_{ k }^{ s }), g_t({ \mathbf{x} }_{ k' }^{ t },{ \mathbf{a} }_{ k' }^{ t })  }) } }},}}
\end{equation}
where $\omega(\cdot, \cdot)$ is a contrastive function that performs semantic alignment or separation in terms of supervised information from both domains. Formally, $\omega(\cdot,\cdot)$ is defined as:
\begin{equation}
\omega(\mathbf{u}, \mathbf{v}) \!=\! \left\{ \begin{array}{ll}
\|\mathbf{u} \!-\! \mathbf{v} \|^{2}             & \text{if} \quad y^s_k = y^t_{k'},\\
\max(0, D \!-\!\|\mathbf{u} \!-\! \mathbf{v} \|^{2})  & \text{if} \quad y^s_k \neq y^t_{k'},
\end{array} \right.
\end{equation}
where $D$ is a parameter dictating the degree of separation and is set to 1 in our experiments. 

\subsection{Alternating Training}
Combining the losses we introduced before together with a $\ell_2$ regularization, we constitute the overall losses for the source and target networks as:
\begin{eqnarray}
{ \mathcal{L} }_{ src } &=& {\mathcal{L}}_{sen}^{s} + \mathcal{L}_{aux} +\lambda  { \mathcal{L} }_{ cfa } + \rho { \mathcal{ L } }_{ reg }^{s}, 
\label{eqn:src_loss}
\\
{ \mathcal{L} }_{ tar } &=& {\mathcal{L}}_{sen}^{t} + \lambda { \mathcal{L} }_{ cfa } +\rho { \mathcal{ L } }_{ reg }^{t}, 
\label{eqn:tgt_loss}
\end{eqnarray}
\noindent
where $\lambda, \rho$ balance the effect of the CFA loss and the $\ell_2$ regularization loss, respectively. The source network has one more auxiliary loss $\mathcal{L}_{aux}$ compared with the target one to achieve task alignment. The whole training procedure consists of two stages: (1) To prevent early overfitting of the target domain, the source network $S$ is individually trained on the source domain by optimizing $\small{{\mathcal{L}}_{sen}^{s}}+ \mathcal{L}_{aux}+ \rho { \mathcal{ L } }_{ reg }^{s}$. Then, $S$ and the BiLSTM, C2A, and PaS modules of $S$ are used to initialize the source and target networks of  the MGAN, respectively. (2) We alternately optimize ${ \mathcal{L} }_{ src }$ for the source network and ${ \mathcal{L} }_{ tar }$ for the target network.

\section{Experiments}

\subsection{Datasets}
\noindent \textbf{Source: AC-level}
We build a large-scale, multi-domain dataset named \emph{{YelpAspect}} as source domains, which is obtained similarly as the Yelp recommendation dataset~\cite{bauman2017aspect}. Specifically, YelpAspect contains three domains: Restaurant (\textbf{R1}), Beautyspa (\textbf{B}), and Hotel (\textbf{H}). The statistics of the YelpAspect dataset are summarized in Table \ref{table1}. Yelp reviews are collected in US cities over six years. Aspect categories and sentiment labels are identified by the ``industrial-strength'' Opinion Parser (OP) system~\cite{qiu2011opinion,liu2015sent}. To be consistent with the target domain datasets, YelpAspect is preprocessed in the sentence level by OP, while the dataset in~\cite{bauman2017aspect} is in the document level. Moreover, we manually double-check to correct wrong annotations produced by OP system and purposely select more negation, contrastive and question instances to make it more challenging. The dataset is available at \url{https://github.com/hsqmlzno1/MGAN}.

\begin{table}[htb]\small
\centering
\resizebox{1.0\linewidth}{!}{
\begin{tabular}{lcclclclc|c|}
  \Xhline{1pt}
  Source domain         &  & {\#}Pos & {\#}Neu & {\#}Neg & {\#}Asp \\ \hline
    \multirow{2}{*}{Restaurant (R1)}  
  & Train & 46,315  & 45,815  & 16,020 &  \multirow{2}{*}{68}  \\ 
  & Test  & 5,207    & 4,944    & 1,743   \\   \hline 
      \multirow{2}{*}{Beautyspa (B)}        
  & Train & 45,770  & 42,580   & 16,023 & \multirow{2}{*}{45} \\ 
  & Test  & 5,056    & 4,793     & 1,823      \\   \hline   
      \multirow{2}{*}{Hotel (H)} 
  & Train & 40,775  & 36,901   & 20,864 & \multirow{2}{*}{44}  \\ 
  & Test &  4,418    & 4,048     & 2,450   \\ 
    \Xhline{1pt}
\end{tabular}}
\caption{The YelpAspect dataset. {\#}Asp denotes the number of aspect categories.}
\label{table1}
\end{table}

\vspace{-1mm}
\noindent \textbf{Target: AT-level} For target domains, we use three public benchmark datasets: Laptop (\textbf{L}), Restaurant (\textbf{R2}) and Twitter (\textbf{T}). The Laptop and Restaurant are from SemEval'14 ABSA challenge ~\cite{kiritchenko2014nrc} by removing a few examples which have ``conflict labels'' as done in~\cite{chen2017recurrent}. The Twitter dataset is collected by \cite{dong2014adaptive}, containing ungrammatical twitter posts. Table \ref{table2} summarizes the statistics of the target domain datasets. 

\begin{table}[htb]\small
\centering
\resizebox{0.8\linewidth}{!}{
\begin{tabular}{lcclclclc|}
  \Xhline{1pt}
  Target Domain &  &{\#}Pos & {\#}Neu & {\#}Neg \\ \hline
  \multirow{2}{*}{Laptop (L)}  
   & Train   & 980  & 454  & 858 \\
   & Test     & 340  & 171  & 128 \\ \hline 
  \multirow{2}{*}{Restaurant (R2)}  
   & Train   & 2,159  & 632  & 800 \\
   & Test    & 730    & 196   & 195 \\ \hline 
  \multirow{2}{*}{Tweets (T)}  
   & Train   & 1,567  & 3,127 & 1,563 \\
   & Test    &  174    & 346   & 174  \\ 
   \Xhline{1pt}
\end{tabular}}
\caption{Statistics of the target domain datasets.}
\vspace{-5mm}
\label{table2}
\end{table}

\subsection{Experimental Setup}
\vspace{-0.5mm}
To evaluate our proposed method, we construct eight coarse-to-fine transfer tasks: R1$\rightarrow$L, H$\rightarrow$L, B$\rightarrow$L, H$\rightarrow$R2, B$\rightarrow$R2, R1$\rightarrow$T, H$\rightarrow$T, B$\rightarrow$T,  where we do not use the pair (R1, R2) as they come from the same domain. For each transfer pair $D_{s}$$\rightarrow$$D_{t}$, the training data from domain $D_{s}$ and randomly sampled 90\% training data from domain $D_{t}$ are used for training, the rest 10\% training data from $D_{t}$ is used for validation, and the testing data from $D_{t}$ is used for testing. Evaluation metrics are Accuracy and Macro-Average F1, where the latter is more suitable for imbalanced datasets.

\vspace{-1mm}
\subsection{Implementation Details}
\vspace{-0.5mm}
The word embeddings are initialized with 200-dimension GloVE vectors~\cite{pennington2014glove} and fine-tuned during the training. $d_{e},d_{h},d_{u}$ are set to be 200, 150 and 100, respectively. The fc layer size is 300. The Adam~\cite{kingma2014adam} is used as the optimizer with the initial learning rate $10^{-4}$. Gradients with the $\ell_2$ norm larger than 40 are normalized to be 40. All weights in networks are randomly initialized from a uniform distribution $\small{U(-0.01,0.01)}$. The batch sizes are 64 and 32 for source and target domains, respectively. The control-off factors $\lambda$, $\rho$ are set to be 0.1 and $10^{-6}$. To alleviate overfitting, we apply dropout on the word embeddings of the context with dropout rate 0.5. We also perform early stopping on the validation set during the training process. The hyperparameters are tuned on 10\% randomly held-out training data of the target domain in R1$\rightarrow$L task and are fixed to be used in all transfer pairs.

\begin{table*}[htb]\small
\centering
\resizebox{0.9\textwidth}{!}{
\begin{tabular}{cclc|cc|ccc}
\Xhline{2\arrayrulewidth}
\multirow{2}{*}{} &  \multirow{2}{*}{\textbf{Model}} & \multicolumn{2}{c}{\textbf{L}} &\multicolumn{2}{c}{\textbf{R2}} &\multicolumn{2}{c}{\textbf{T}} \\ \cline{3-8} 
\multicolumn{2}{c}{} 				    & Acc  & Macro-F1  & Acc  & Macro-F1 & Acc  & Macro-F1    \\
\hline 
\multirow{6}{*}{\textbf{Baselines}} 
					            &AE-LSTM~\cite{wang2016attention}                  & 68.97 & 62.50  & 76.25 & 64.32 & 69.42 & 56.28        \\
					            &ATAE-LSTM~\cite{wang2016attention}              & 68.65 & 62.45  & 77.23 & 64.95 & 69.58 & 56.72        \\
					            &TD-LSTM~\cite{tang2015effective}                  & 68.18 & 62.28  & 75.63 & 64.16 &66.62 & 64.01         \\
					            &IAN~\cite{ma2017interactive}                           & 72.10 & -          & 78.60 & -        & -        & -        	        \\
                                &MemNet~\cite{tang2016aspect}                    & 70.33 & 64.09  & 78.16 & 65.83 &68.50 & 66.91         \\
                                &RAM~\cite{chen2017recurrent}                         & 72.08 & 68.43  & 78.48 & 68.54 &69.36 & 67.30          \\
\hline 
\multirow{1}{*}{\textbf{Base Model}} 
                                                      &TN		              & 70.58 & 65.34  & 77.91 & 65.75 &71.68 & 71.02 	 \\
\hline 

\multirow{2}{*}{} &  \multirow{2}{*}{} & \multicolumn{6}{c}{\textbf{Average results over each target domain}} \\ 
\hline  
\multirow{2}{*}{\textbf{Ablated Models}} 
                                                      &MGAN w/o PI            & 72.98 & 67.71  & 78.99 & 66.41  & 72.88 & 71.57  \\
                                                      &MGAN w/o C2F        & 74.80 & 69.63  & 80.46 & 67.86  & 73.53 & 72.37 \\
\hline
\multirow{1}{*}{\textbf{Full Model}} 
                                                      &MGAN                      & \textbf{76.21}$^{\dag}$$^{\ddag}$ & \textbf{71.42}$^{\dag}$$^{\ddag}$  & \textbf{81.49}$^{\dag}$$^{\ddag}$ & \textbf{71.48}$^{\dag}$$^{\ddag}$ &\textbf{74.62}$^{\dag}$$^{\ddag}$ & \textbf{73.53}$^{\dag}$$^{\ddag}$ \\
\Xhline{1.5\arrayrulewidth}
\end{tabular}}
\vspace{-2mm}
\caption{Experimental results (\%). The marker $^{\dag}$ refers to $p$-value $<$ 0.05 when comparing with MGAN w/o C2F , while the marker $^{\ddag}$ refers to $p$-value $<$ 0.05 when comparing with RAM. }
\vspace{-4mm}
\label{table:mr1}
\end{table*}

\vspace{-2mm}
\subsection{Baseline Methods}
The baseline methods are divided into two groups:\\
\noindent \textbf{Non-Transfer} To demonstrate the benefits from coarse-to-fine task transfer, we compare with the following state-of-the-art AT-level methods without transfer: 
\begin{itemize}
\item \textbf{TD-LSTM}~\cite{tang2015effective}: It employs two LSTMs to model the left and right contexts of the aspect and then concatenates the context representations for prediction.
\item \textbf{AE-LSTM}, and \textbf{ATAE-LSTM}~\cite{wang2016attention}: AE-LSTM is a simple LSTM model incorporating the aspect embedding as input, while ATAE-LSTM extends AE-LSTM with the attention mechanism.
\item \textbf{MemNet}~\cite{tang2016aspect}: it applies a memory network with multi-hops attentions and predicts sentiment based on the top-most context representations.

\item \textbf{IAN}~\cite{ma2017interactive}: It adopts two LSTMs to learn the representations of the context and the aspect interactively; 
\item \textbf{RAM}~\cite{chen2017recurrent}: It employs multiple attentions with a GRU cell to non-linearly combine the aggregation of word features in each layer. 
\item \textbf{Target Network (TN)}: It is our proposed base model (BiLSTM+C2A+Pas) trained on ${D}_{t}$ for the target task.
\end{itemize}
For IAN, we report the results in the original paper and use the source codes of other methods for experiments. \vspace{-2mm}
\\\\
\noindent \textbf{Transfer} To investigate the effectiveness of the CFA , we also compare the following transfer methods:
\begin{itemize}
\item \textbf{Source-only (SO)}: It uses a source network trained on ${D}_{s}$ to initialize a target network and then tests it on ${D}_{t}$.
\item \textbf{Fine-tuning (FT)}: It advances SO with further fine-tuning the target network on ${D}_{t}$.
\item \textbf{M-DAN}: It is a multi-adversarial version of Domain Adversarial Network (DAN)~\cite{ganin2016domain} based on multiple domain discriminators. All discriminators are built upon the PaS layers of the two networks, each of which aligns one class distribution between domains.
\item \textbf{M-MMD}: Similar with M-DAN, M-MMD aligns different class distributions between domains based on multiple Maximum Mean Discrepancy (MMD)~\cite{gretton2007kernel}. For each MMD, 
following the~\cite{bousmalis2016domain}, 
we use a linear combination of 19 RBF kernels with the width parameters ranging from $10^{-6}$ to $10^{6}$.
\end{itemize}
The original DAN and MMD are unsupervised domain adaptation methods. Thus, for fair comparison, we use the source code of DAN and MMD, and extend them to M-DAN and M-MMD that utilize target supervised information and have higher performances, respectively.

\subsection{Result Analysis}
\noindent \textbf{Comparison with Non-Transfer}
Note that we are the first to explore transfer techniques and achieve the best performances in this task. Thus, it is necessary to show our improvements over current superior non-transfer methods. The classification results are shown in Table~\ref{table:mr1}. The results of our full model and its ablations are calculated by averaging over each target domain among eight transfer pairs (e.g., R2 is obtained by averaging over H$\rightarrow$R2 and B$\rightarrow$R2). Based on the table, we have the following observations: (1) Our full model MGAN consistently and significantly achieves the best results in all target domains, outperforming the strongest baseline RAM by 4.13\%, 3.58\%, 5.26\% for accuracy and 2.99\%, 2.94\% and 6.23\% for Macro-F1 on average. Our base model TN that does not utilize the knowledge from the source task, can only compete against with the baselines. It could be more convincing that the MGAN can achieve superior performances even with a simple model for the target task. This also indicates that the efficacy of MGAN benefits from leveraging useful knowledge learned from the source task. (2) MGAN consistently outperforms the MGAN w/o C2F, where C2F module of the source network is removed and the source position information is missed (we set all ${ p }^{ s }_{i}$ to 1), by 1.41\%, 1.03\%, 1.09\% for accuracy and 1.79\%, 3.62\% and 1.16\% for Macro-F1 on average. This is because that the C2F can effectively reduce the aspect granularity gap between tasks such that more useful knowledge can be distilled to facilitate the target task. (3) Position information is crucial for aspect-level sentiment analysis. The MGAN w/o PI, which does not utilize the position information, performs very poorly.

\begin{table}[tb]\small
\centering
\resizebox{1\linewidth}{!}{
\begin{tabular}{c|cccccc}
\Xhline{2.5\arrayrulewidth}
Acc  & \multirow{2}{*}{SO} & \multirow{2}{*}{FT} & \multirow{2}{*}{M-DAN} & \multirow{2}{*}{M-MMD}  & \multirow{2}{*}{MGAN w/o SS} & \multirow{2}{*}{MGAN}   \\
Macro-F1  \\ 
\Xhline{2.5\arrayrulewidth}

\multirow{2}{*}{R1$\rightarrow$L}   
& 69.80	& 74.80	& 75.74	& 75.90	 & 77.00	   & \textbf{77.62}				   \\
& 67.05    & 69.84     & 71.13 	& 70.95	 & 71.31	   & \textbf{72.26}			   \\ 
\hline
\multirow{2}{*}{B$\rightarrow$L}   
&70.27  	& 71.99	& 72.46	& 74.02	& 74.49 	   & \textbf{75.74}			  \\ 
&66.84   	& 67.13 	& 68.69	& 69.36	& 69.94  	   & \textbf{71.65}			 \\
\hline
\multirow{2}{*}{H$\rightarrow$L}   
&70.74     	& 72.77	& 75.43	& 73.71	& 74.02	    & \textbf{75.27}	 \\
&67.89     & 67.75     & 71.40 	& 69.16	& 69.31   	    & \textbf{70.34}		  \\
\hline

\multirow{2}{*}{B$\rightarrow$R2}   
&72.90	& 79.16	& 79.96	& 81.31	& \textbf{81.84}	   & 81.66			\\ 
&64.36    	& 66.78	& 68.73	& 70.54	& \textbf{71.80}       & 71.72				\\
\hline
\multirow{2}{*}{H$\rightarrow$R2}   
&72.36 	& 80.59   	& 79.87 	& 79.87	& 80.95 	    & \textbf{81.31}				 \\
&62.48	& 69.57   	& 69.19 	& 67.58	& 70.57    	    & \textbf{71.24}				 \\
\hline

\multirow{2}{*}{R1$\rightarrow$T}   
&46.39 	& 72.83 	& 72.11	& 73.41	& 73.41	   & \textbf{75.00}					\\ 
&45.74      & 72.10 	& 70.69	& 72.52	& 72.76	   & \textbf{74.00}					   \\
\hline
\multirow{2}{*}{B$\rightarrow$T}   
&46.39   	& 72.25	& 72.98 	& 73.27	& 73.27 	   & \textbf{74.00}						 \\
&45.62 	& 70.30	& 71.88	& 72.34	& 71.79	   & \textbf{72.87}						   \\
\hline
\multirow{2}{*}{H$\rightarrow$T}   
&47.40 	& 71.82	& 72.55	& 73.27	 & 73.99        & \textbf{74.86}						 \\ 
&46.71  	& 70.05	& 71.07	& 72.11	 & 72.32	    & \textbf{73.73}						\\

\Xhline{2.5\arrayrulewidth}
\multirow{2}{*}{Average}   
&62.03	& 74.53	& 75.13	& 75.60	& 76.12 	     & \textbf{76.93}$^{\dag}$					 \\ 
&58.34	& 69.19	& 70.33	& 70.	57	& 71.23         & \textbf{72.23}$^{\dag}$					\\
\Xhline{2.5\arrayrulewidth}
\end{tabular}}
\vspace{-2mm}
\caption{Experimental results (\%). The marker $^{\dag}$ refers to $p$-value $<$ 0.05 when comparing with MGAN w/o SS.}
\label{table:mr2}
\vspace{-6mm}
\end{table}

\begin{figure*}[tb!]
\centering
\subfigure{\includegraphics[width=0.77\textwidth]{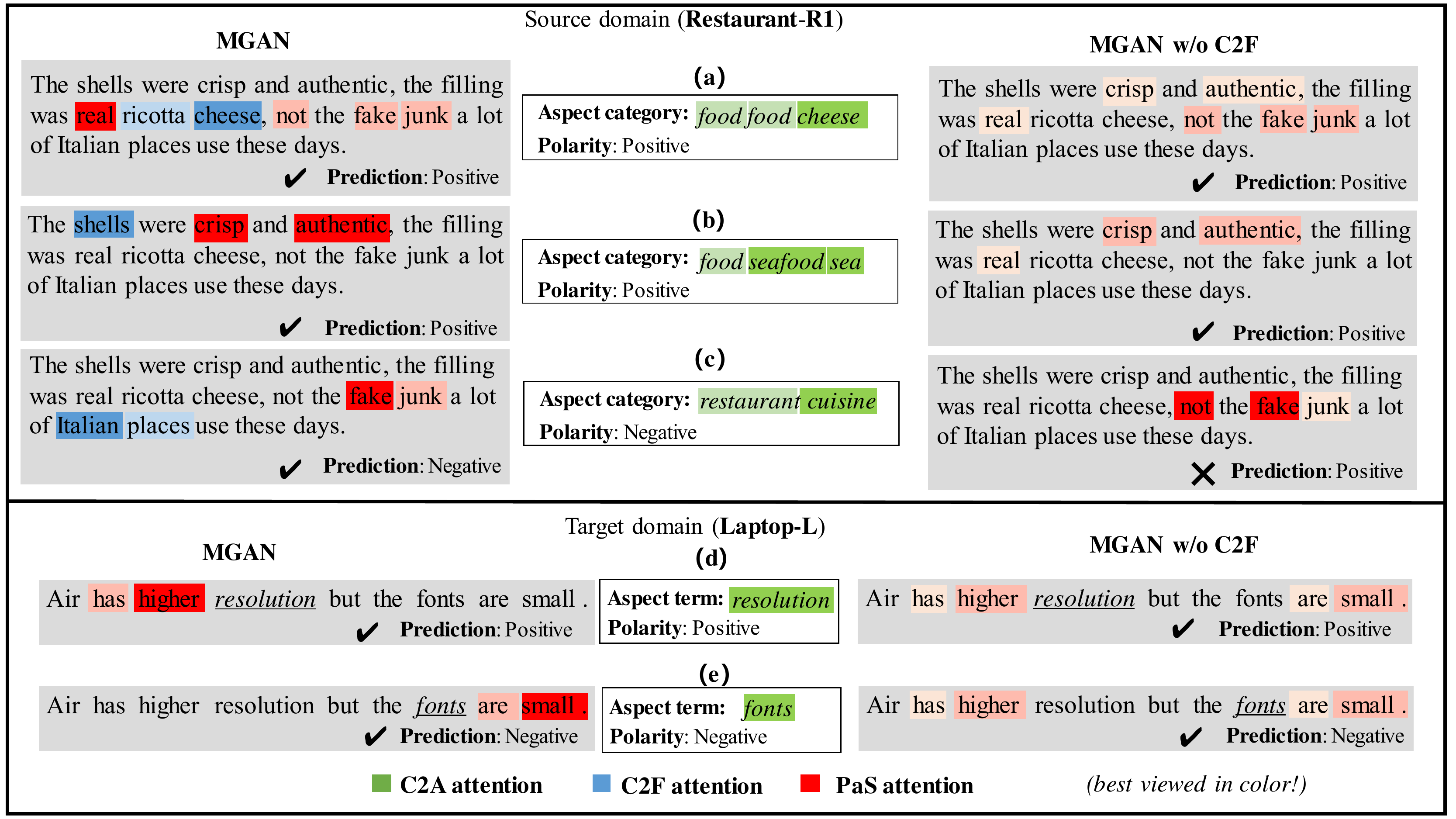}}
\vspace{-4mm}
\caption{Visualization of attention: MGAN versus MGAN w/o C2F in the R1$\rightarrow$L task. Deeper color denotes higher weights.}
\vspace{-6mm}
\label{attention}
\end{figure*}

\vspace{1mm}
\noindent \textbf{Comparison with Transfer} To avoid the effect of aspect granularity gap, all these models keep the C2F module. The compared results are shown in Table~\ref{table:mr2}. SO performs poorly due to no adaptation applied. The popular technique FT cannot achieve satisfactory results since fine-tuning may cause the oblivion of useful knowledge from the source task. The full model MGAN outperforms M-DAN and M-MMD by 1.80\% and 1.33\% for accuracy and 1.90\% and 1.66\% for Marco-F1 on average, respectively. We derive two possible reasons: First, enormous target data is unavailable since it is hard to obtain, thus, it may be insufficient to represent target distributions by limited target labeled data for the distribution alignment based methods; Second, M-DAN and M-MMD focus on the semantic alignment but ignore semantic separation. Remarkably, MGAN considers both of them in a point-wise surrogate, which altogether improves the performance of our method. Besides, MGAN outperforms its ablation MGAN w/o SS removing the semantic separation loss of the CFA by 0.81\% for accuracy and 1.00\% for Macro-F1 on average, which implies that the semantic separation plays an important role in alleviating false alignment.

\begin{figure}[tb!]
\centering
\includegraphics[width=0.89\linewidth]{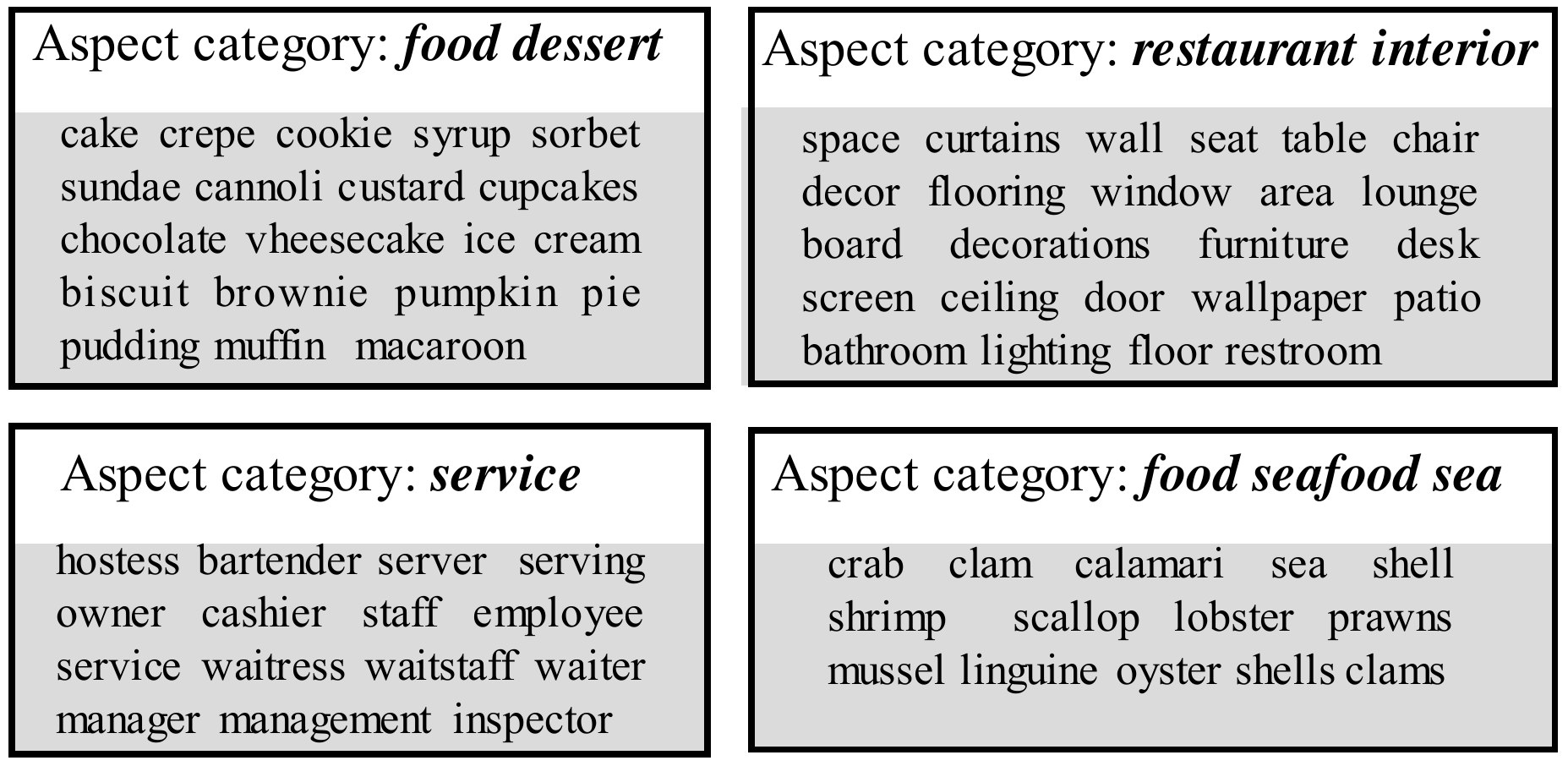}
\vspace{-2mm}
\caption{Associated aspect terms towards different aspect categories captured by C2F attention in the R1$\rightarrow$L task.}
\vspace{-5mm}
\label{fig:words}
\end{figure}

\vspace{-1mm}
\subsection{Effect of C2F Attention Module}
\vspace{-0.5mm}
We now give some illustrated examples to show the effect of C2F for solving aspect granularity inconsistency, by comparing MGAN and MGAN w/o C2F. Some hard cases containing multiple sentiment-aspect paris in the R1$\rightarrow$L task are shown in Figure~\ref{attention}. In the source domain R1, both models first utilize the C2A to attend the informative part of the aspect category, e.g., {\it ``cheese''},  {\it ``seafood sea''} and {\it ``cuisine''}, which are representatives for each aspect. Then, compared with MGAN w/o C2F, MGAN further uses C2F to capture more specific aspect terms from the context towards the aspect category, such as {``shells''} to {\it food seafood sea}, which helps the source task capture more fine-grained semantics of aspect category and detailed position information like the target task, such that the sentiment attention can be position-aware and identify more relevant sentiment features towards the aspect. For example, in the (a) and (c), the user expresses a positive sentiment over {\it food food cheese} but a negative attitude towards {\it restaurant cuisine} ({\it cuisine} means a style of cooking especially as a characteristic of a particular country or region). MGAN captures the regional words for the cooking style, i.e., {``italian place''} towards {\it restaurant cuisine} and the related n-gram sentiment feature {``fake junk''} instead of the {``not the fake junk''} for the {``ricotta cheese''}, and finally makes a correct prediction, which helps distill more useful knowledge for subsequent feature alignment. While MGAN w/o C2F locates wrong sentiment contexts and fails in (c). As such, benefited from distilled knowledge from the source task, MGAN can better model the complicated relatedness between the context and aspect term for the target domain L, but MGAN w/o C2F performs poorly though it make true predictions in (d) and (e). Moreover, as shown in Figure \ref{fig:words}, we list some samples of captured associated aspect terms towards different aspect categories based on the highest C2F attention weight. These underlying aspect terms make the source task more correlated to the target task.

\vspace{-2mm}
\section{Conclusion and Future Work}
\vspace{-0.5mm}
In this paper, we explore a motivated direction for aspect-level sentiment classification named {\it coarse-to-fine task transfer} and build a large-scale YelpAspect dataset as highly beneficial source benchmarks. A novel MGAN model is proposed to solve both aspect granularity inconsistency and domain feature mismatch problems, and achieves superior performances. Moreover, there are many other potential directions, like transferring between different aspect categories across domains, transferring to a AT-level task where the aspect terms are also not given and need to be firstly identified. We believe all these can help improve the ASC task and there will be more effective solutions coming in the near future.

\section{Acknowledgement}
This work is supported by Hong Kong CERG grants (16209715 and 16244616), and NSFC (61473087 and 61673202).

\small
\bibliography{MGAN}
\bibliographystyle{aaai}

\end{document}